\documentclass[acmlarge]{acmart}
\pdfoutput=1
\makeatletter
\renewcommand\@formatdoi[1]{\ignorespaces}
\makeatother

\setcopyright{none}
\renewcommand\footnotetextcopyrightpermission[1]{} % removes footnote with conference information in first column

\AtBeginDocument{%
  }
\setcopyright{acmcopyright}
\copyrightyear{2023}
\acmConference[ICMLT 2023]{2023 8th International Conference on Machine Learning Technologies (ICMLT)}{March 10--12 2023, Stockholm, Sweden}
\usepackage{hyperref}
\usepackage{graphicx}
\usepackage{float}
\usepackage[linesnumbered,ruled]{algorithm2e}

\usepackage{tikz}
\usepackage{tikzsymbols}
\usetikzlibrary{positioning}
\usetikzlibrary{calc,through}
\usepackage{multirow}
\usepackage{booktabs} % tabluar: toprule, midrule, ...
\usepackage{amsmath}
\usepackage{arydshln} % dashed horizontal line in table
\usepackage{color, colortbl} % colors and table coloring
\usepackage[htt]{hyphenat}
\usepackage{fontawesome}
\usepackage{listings}
\usepackage{enumitem}
\usepackage{wrapfig}
\usepackage{caption}
\usepackage{subcaption}

\usepackage{pgfplots}
\pgfplotsset{
    compat=1.17,
}
\usepackage{pgfplotstable}

\newcolumntype{L}[1]{>{\raggedright\let\newline\\\arraybackslash\hspace{0pt}}m{#1}}
\newcolumntype{C}[1]{>{\centering\let\newline\\\arraybackslash\hspace{0pt}}m{#1}}
\newcolumntype{R}[1]{>{\raggedleft\let\newline\\\arraybackslash\hspace{0pt}}m{#1}}

\begin{document}
\pagestyle{plain}
% \title{HE-MAN -- Homomorphically~Encrypted MAchine~learning~with~oNnx~models}
\title{On the Effect of Adversarial Training Against Invariance-based Adversarial Examples}

\author{Roland Rauter}
\affiliation{%
  \institution{MCI~Management~Center~Innsbruck, Innsbruck}
  \city{Innsbruck}
  \country{Austria}
  \postcode{6020}
}
%\email{rr9376@mci4me.at}

\author{Martin~Nocker}
\authornote{Correspondence: martin.nocker@mci4me.at, Tel.: +43-512-2070-4300, MCI~Management~Center~Innsbruck, Universitätsstraße 15, 6020 Innsbruck, Austria}
\affiliation{%
  \institution{MCI~Management~Center~Innsbruck, Innsbruck}
  \city{Innsbruck}
  \country{Austria}
  \postcode{6020}
}
%\email{martin.nocker@mci.edu}

\author{Florian~Merkle}
\affiliation{%
  \institution{MCI~Management~Center~Innsbruck, Innsbruck}
  \city{Innsbruck}
  \country{Austria}
  \postcode{6020}
}

\author{Pascal~Schöttle}
\affiliation{%
  \institution{MCI~Management~Center~Innsbruck, Innsbruck}
  \city{Innsbruck}
  \country{Austria}
  \postcode{6020}
}
%\email{pascal.schoettle@mci.edu}

\renewcommand{\shortauthors}{Roland Rauter et al.}

\begin{abstract}
Adversarial examples are carefully crafted attack points that are supposed to fool machine learning classifiers. In the last years, the field of adversarial machine learning, especially the study of perturbation-based adversarial examples, in which a perturbation that is not perceptible for humans is added to the images, has been studied extensively. Adversarial training can be used to achieve robustness against such inputs. Another type of adversarial examples are invariance-based adversarial examples, where the images are semantically modified such that the predicted class of the model does not change, but the class that is determined by humans does. How to ensure robustness against this type of adversarial examples has not been explored yet. This work addresses the impact of adversarial training with invariance-based adversarial examples on a convolutional neural network (CNN).

We show that when adversarial training with invariance-based and perturbation-based adversarial examples is applied, it should be conducted simultaneously and not consecutively. This procedure can achieve relatively high robustness against both types of adversarial examples. 
Additionally, we find that the algorithm used for generating invariance-based adversarial examples in prior work does not correctly determine the labels and therefore we use human-determined labels.
%However, it remains to be mentioned that there are high fluctuations in these accuracies. Thus, it cannot be guaranteed that a model is robust to both types of adversarial examples.
\end{abstract}




\maketitle
\begin{figure}
    \begin{subfigure}[H]{0.2\textwidth}
        \includegraphics[width=\textwidth]{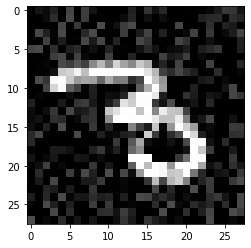}
         \caption{}
        \label{fig:ptb-example}
    \end{subfigure}
    \quad
    \begin{subfigure}[H]{0.2\textwidth}
        \includegraphics[width=\textwidth]{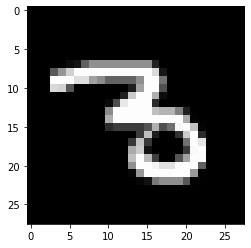}
         \caption{}
        \label{fig:original-example}
    \end{subfigure}
    \quad
    \begin{subfigure}[H]{0.2\textwidth}
        \includegraphics[width=\textwidth]{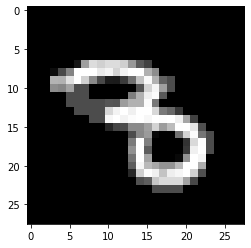}
         \caption{}
        \label{fig:inv-example}
    \end{subfigure}
    \caption{\ref{fig:original-example} shows the original image taken from the MNIST dataset. 
     \ref{fig:ptb-example} represents a perturbation-based adversarial example based on the original image. The image still displays the number three for humans but the model would classify this as a number five. 
    \ref{fig:inv-example} represents an invariance-based adversarial example based on the original image. The image that displays the number three now displays the number eight as seen by humans. The model would classify this still as a number three.}
    \label{fig:example}
\end{figure}

\section{Introduction}
Machine learning (ML) and deep learning (DL)-based algorithms invade more and more areas of our daily lives. To name just a few examples, medicine~\cite{ker2017deep}, cyber security~\cite{alazab2019deep}, and finance~\cite{huang2020deep} all benefit from the usage of deep learning-based algorithms. Naturally, such application areas invoke adversaries that have incentives to attack ML-based algorithms and since the seminal work of Szegedy et al.~\cite{szegedy2013intriguing} that coined the term ``adversarial examples'' for these attack points, such attacks have been developed for several real-world scenarios like road sign classification~\cite{eykholt2018robust} for autonomous driving, fooling voice-assistants~\cite{carlini2018audio}, or face-recognition software with adversarial eyeglass frames~\cite{sharif2016accessorize}.

%%% moved here from Section 2
Adversarial examples can be divided into two types: Firstly, perturbation-based adversarial examples, where a perturbation that is commonly not perceptible to human eyes is added to an image. When such an adversarial example is successful, the model misclassifies the image, while humans still assign the original label. Exemplarily, if an image shows the digit three (Figure~\ref{fig:original-example}) and a successful perturbation-based adversarial example is created (Figure~\ref{fig:ptb-example}) the model no longer classifies this as the digit three, while (most) humans still do. %Overall, a perturbation-based adversarial example is successful when the classification of the ML classifier changes but the classification of humans does not.

In contrast, invariance-based adversarial examples~\cite{jacobsen2019exploiting} change the image semantically, such that it looks different for humans than the original image and they classify it differently from the original image. However, with this type of adversarial examples, the model does not change its classification. For example, an image that shows the number three (Figure~\ref{fig:original-example}) could be modified to look like an eight for humans, as in Figure~\ref{fig:inv-example}. In this case, the model still classifies the image as a three. Figure~\ref{fig:example} compares both types of adversarial examples.

%%%%

In this paper, we present the effects of adversarial training using invariance-based adversarial examples. % has on a CNN that is trained with the MNIST dataset. 
To be precise, accuracy and robustness against both types of adversarial examples of a convolutional neural network (CNN) that is trained with the MNIST dataset are analyzed. % and exhibits an accuracy of 99\,\% on the MNIST test data and an accuracy of 0\,\% on perturbation-based adversarial examples. 
The same techniques are applied to a CNN with identical architecture that is already trained with perturbation-based adversarial examples.
% Moreover, we evaluate a CNN with identical architecture that is already trained with perturbation-based adversarial examples. % and exhibits an accuracy of 96.2\,\% on the MNIST test data and an accuracy of 88.9\,\% on perturbation-based adversarial examples. 

We investigate if robustness against adversarial examples of both types increases if we include invariance-based adversarial examples during perturbation-based adversarial training in two different ways: performing perturbation-based and invariance-based adversarial training subsequently and secondly, adversarial training using both types of adversarial examples simultaneously.
We identify a tradeoff between robustness against perturbation-based and invariance-based adversarial examples, depending on the training procedure.
Moreover, we find sweet spots of robustness with respect to the ratio between perturbation-based and invariance-based adversarial examples.

The perturbation-based adversarial examples are generated using the $l_\infty$-projected gradient descent~\cite{madry2018towards}. Invariance-based adversarial examples are generated using the algorithm proposed by Tramer et al.~\cite{pmlr-v119-tramer20a}. The labels of invariance-based adversarial examples have to be defined explicitly when performing training. Because of that, we investigate both, human-defined labels and algorithm-defined labels. 

~\\
To summarize, we make the following contributions:
\begin{itemize}[noitemsep,topsep=0pt]
    % \item We analyze the effects of adversarial training using invariance-based adversarial examples on:
    % \begin{itemize}
    %     \item a CNN that is not adversarially-trained with perturbation-based adversarial examples, and
    %     \item a CNN that is already adversarially-trained with perturbation-based adversarial examples.
    % \end{itemize}
    \item We analyze the effects of adversarial training using invariance-based adversarial examples on a CNN that is trained using benign data, and a model that is additionally trained using perturbation-based adversarial examples.
    \item We examine the model performance when using invariance-based adversarial examples during training with labels generated by the algorithm from~\cite{pmlr-v119-tramer20a} compared to human-defined labels.
    \item We further evaluate the model robustness depending on whether perturbation-based and invariance-based adversarial training is performed sequentially or simultaneously.
    \item We identify sweet spots of robustness with respect to the ratio between perturbation-based and invariance-based adversarial examples.
\end{itemize}
~\\
The rest of this paper is organized as follows. First, section~\ref{sec:related-work} presents related work, before sections~\ref{sec:experiments} and~\ref{sec:results} discuss the approach of consecutive retraining and the corresponding results. Section~\ref{sec:simultaneous} introduces an improved approach of simultaneous adversarial training and its results. Finally, section~\ref{sec:discussion} discusses the results and limitations of our investigations before section~\ref{sec:conclusion} concludes this paper.

% Not sure if we need both, RW and BG...
\section{Related Work}\label{sec:related-work}
%\textbf{TODO} describe epsilon too?\\\\ DONE
% The field of Adversarial Examples is divided into two types. First, there are perturbation-based adversarial examples where a perturbation to an image is added that is not or only partly visible. When this is done correctly, a CNN classifies this image wrong. For example, if there is an image that shows the number two and a Perturbation-Based Example is created out of it, the CNN classifies this as not a number of two. In short, a Perturbation-Based Adversarial Example is successful when the classification of the CNN changes but not the classification of us humans.\\
% In contrast to that, there are invariance-based adversarial examples where the image is semantically changes so that it changes and for us human it looks different than the original image and we classify it different than the original image. The CNN does not change the classification of the image. For example, an image that shows the number one. It is changed that is looks like a seven for humans. The CNN states that this image still shows the number one.\\

Research on \emph{adversarial machine learning} started in 2004 when it was first explored that spam filters utilizing linear classifiers can be fooled by small changes in the initial email that do not negatively affect the readability of the message but lead to misclassification~\cite{dalvi2004adversarial}. 
In 2013, Szegedy et al.~\cite{szegedy2013intriguing} showed that deep neural networks (DNNs) are just as prone to adversarial examples as other machine learning algorithms, when classifying carefully perturbed input samples.

Formally, a \emph{perturbation-based} adversarial example can be described as follows: A classifier is a function $x\mapsto C(x)$ that takes an input $x$ and yields a class $C(x) = y$. If a semantic-preserving perturbation $\delta$ is added to $x$, such that the manipulated input $x + \delta = \tilde{x}$ leads to a classification different from the original value $C(\tilde{x}) \neq C(x)$, it is labeled an adversarial example.
Usually, distance metrics are used to quantify the difference between $x$ and $\tilde{x}$ and it is enforced that $||x - \tilde{x}|| \leq \epsilon$ for a given value of $\epsilon$.

Several methods to make ML classifiers robust against perturbation-based adversarial examples exist. The most promising of them is \emph{adversarial training}~\cite{goodfellow2015explaining,madry2018towards}, where the training dataset is extended by perturbation-based adversarial examples. This results in an improved robustness against those attacks. Other countermeasures exist, e.g.~\cite{goodfellow2015explaining,metzen2017detecting,pang2018robust} that try to identify perturbation-based adversarial examples at inference time, but they do not work reliable enough to call ML classifiers robust against these attacks.

Further, \cite{jacobsen2019exploiting} introduced the term \emph{invariance-based} adversarial examples that exploit the invariance of a model by changing a sample's semantic meaning while preserving the model's classification.
To formally differentiate perturbation-based from invariance-based adversarial examples, we draw on the notion introduced by~\cite{pmlr-v119-tramer20a} having a \textit{label oracle} $\mathcal{O}$ that maps all inputs to their real class, i.e. $\mathcal{O}(x) = y$. This notion helps to further specify a perturbation-based adversarial example as:
\begin{itemize}[noitemsep,topsep=2pt]
    \item[-] $C(\tilde{x}) \neq y$, i.e. the model assigns a different class to the perturbed image% than its original label.
    \item[-] $\mathcal{O}(\tilde{x}) = y$, i.e. the label oracle assigns the true class to the perturbed image
\end{itemize}
whereas for any invariance-based adversarial example it holds that:
\begin{itemize}[noitemsep,topsep=2pt]
    \item[-] $C(\tilde{x}) = y$, i.e. the model assigns the original label to the perturbed image.
    \item[-] $\mathcal{O}(\tilde{x}) \neq y$, i.e. the label oracle assigns the new semantically correct label to the perturbed image.
\end{itemize}
\noindent{}\cite{pmlr-v119-tramer20a} and \cite{jacobsen2019exploiting}, both propose an algorithm for the computational creation of invariance-based adversarial examples. \cite{pmlr-v119-tramer20a} further provide 100 invariance-based adversarial examples, that were crafted by humans with the help of a pixel editor. The pixel editor itself is not openly accessible.

At the point of writing, no approach exists to improve robustness against invariance-based adversarial examples, while preserving robustness against perturbation-based adversarial examples. \cite{jacobsen2020excessive} propose a loss function based on the cross-entropy loss with an added maximum likelihood term to tackle the problem, but do not report the robustness against perturbation-based adversarial examples.

\section{Experiments}\label{sec:experiments}
We analyze the effect of adversarial training using invariance-based adversarial examples with a CNN trained on the MNIST dataset, which is referred to as the ``standard-trained model''. Furthermore, a second model with identical architecture as the standard-trained model and additionally trained using perturbation-based adversarial examples is analyzed. This model is referred to as the ``ptb-trained model''. We use these models as the base models and re-train them as is described in this section.
%In the next two sections, we present the approach of the adversarial training against both types.

\subsection{Data and Model}
The MNIST dataset~\cite{lecun2010mnist} is a dataset of handwritten digits, consisting of 60\,000 training images and 10\,000 test samples. Each image is 28\,$\times$\,28 pixels, with a single digit (0-9) written in grayscale. The task of classifying the digits is a common benchmark for machine learning algorithms, especially in the field of computer vision and image processing. The dataset is widely used as a benchmark for training and testing models in the field of machine learning, particularly in the realm of image recognition and deep learning.

The CNN consists of two convolutional layers with 32 and 64 filters, respectively, with a kernel size of 5\,$\times$\,5 and ReLU activation function. After both convolutional layers, a max-pooling layer with a 2\,$\times$\,2 filter size and a stride of one is applied. After the convolutional layers, the CNN exhibits a fully-connected layer with 1\,024 hidden units and ReLU activation function. The last layer consists of a fully-connected layer with ten output units and a softmax activation function. This is the same architecture as Madry et al.~\cite{madry2018towards} use for their perturbation-based adversarial training.

We train this model by using the RMSprop optimizer~\cite{hinton2012neural}, a batch size of 1024, a learning rate of 0.001, and a maximum number of epochs of 1000. We implement an early stopping criterion, i.e. as soon as the loss with respect to the validation data is increasing, we stop the training process. For calculating the loss, we use the categorical cross-entropy method. This results in an accuracy on the MNIST test data of 99.9\,\%.
All 60\,000 training samples are used for the training process.

\begin{wrapfigure}{r}{0pt}
\begin{minipage}{0.4\textwidth}
\vspace{-4mm}
\begin{algorithm}[H]
\caption{Perturbation-based\\adversarial training}\label{alg:ptb-training}
\KwIn{model $M$, target accuracy $acc_t$, maximum iterations $i_{max}$}
\KwResult{ptb-trained model $M_{ptb}$}
$acc \gets 0$\\
$i \gets 0$\\
\While{$acc \leq acc_t$ \textbf{or} $i\leq i_\text{max}$}{
    $batch \gets next\_batch(1000)$\\
    $adv\_ex \gets attack(M, batch)$\\
    $train(M, adv\_ex)$\\
    $acc \gets test(M)$\\
    $i \gets i + 1$
}
$M_{ptb} \gets M$
\end{algorithm}
\end{minipage}
\end{wrapfigure}
\subsection{Perturbation-based adversarial training}
Generating perturbation-based adversarial examples is an iterative white-box approach in this work. The parameters of the model are updated during every training iteration, and so the perturbation-based adversarial examples are also changing in every iteration. To achieve robustness, repeating the perturbation-based adversarial training is mandatory. For this approach, we apply the algorithm described in the pseudocode of Algorithm~\ref{alg:ptb-training}. The algorithm takes a model, the target accuracy, and the number of training iterations as inputs and results in a ptb-robust model $M_{ptb}$. 
%If the user passes a high accuracy and a high iteration count to the algorithm, it is executed until the iteration count is reached, and vice versa. 
The $next\_batch(n)$ function iterates over the whole MNIST training dataset and returns $n$ data samples, which is fixed to 1000 in our experiments. Next, $attack(M,batch)$ generates perturbation-based adversarial examples before $train(M, adv\_ex)$ retrains the model with the generated samples. For training, the original labels of the MNIST dataset are used. The method $test(M)$ returns the accuracy on perturbation-based adversarial examples. 

To create perturbation-based adversarial examples, we use the Foolbox\footnote{we use version 3.3.1}~\cite{rauber2017foolbox,rauber2017foolboxnative} implementation of the $l_\infty$-projected gradient descent attack~\cite{madry2018towards}. The maximum perturbation $\epsilon$ is set to $0.3$.
For every iteration in the perturbation-based adversarial training process, we generate 1000 perturbation-based adversarial examples based on the 60\,000 MNIST training samples.% After every iteration, we choose the next 1\,000 and so on. Until we reach the last example. Then we start again with the first 1\,000.

% Depending on the choice between iterations or the needed accuracy it is checked against $accuracy\_to\_achieve$ or $max\_iterations$.

\subsection{Invariance-based adversarial training}
Invariance-based adversarial examples are independent of the model, therefore no iterative algorithm for generating these samples is necessary, i.e. this is a black-box approach. %We used the algorithm provided by Tramer et al.~\cite{pmlr-v119-tramer20a} to generate invariance-based adversarial examples in our experiments, before using them for adversarial training in a second step.
%Additionally, as the order in which adversarial training is conducted might matter, we examine whether performing the types of adversarial training simultaneously is favorable, or perform it consecutively.
To create the invariance-based adversarial examples for training the CNN, we use the algorithm provided by Tramer et al.~\cite{pmlr-v119-tramer20a}. For testing we use the human-crafted invariance-based adversarial examples also provided by Tramer et al. \cite{pmlr-v119-tramer20a}.

As there might be a difference in which labels are used when performing invariance-based adversarial training, we analyze two different label assignment approaches. We differentiate between the human-defined labels and the labels defined by the algorithm, i.e. the label of the starting input. To derive the human-defined labels, ten human participants were asked to label 500 invariance-based adversarial examples that were crafted algorithmically. The most selected label of each example is then set as the human label. In total, we use 500 samples with human labels and 500 samples with labels defined by the algorithm. These are then used in the experiments to evaluate the effects when performing adversarial training. This leaves us with a total of four configurations: standard-trained base model re-trained with invariance-based adversarial examples containing human-defined labels (i) and algorithm-defined labels (ii), and ptb-trained base model re-trained with invariance-based adversarial examples containing human-defined labels (iii) and algorithm-defined labels (iv).

\subsection{Evaluation metrics}
As stated in the beginning of this section, we have a standard-trained model and a ptb-trained model. We perform the invariance-based adversarial training on both models. The evaluation metrics are the accuracy and robustness values of the models.
We measure the models' accuracy on the benign MNIST test data. This is referred to as the clean accuracy. In addition to that, we measure the models' accuracy on the perturbation-based adversarial examples. This is referred to as the ptb-robustness. Finally we report the models' accuracy on the invariance-based adversarial examples %(crafted by humans) provided by Tramer et al.~\cite{pmlr-v119-tramer20a}. This is ,
referred to as the inv-robustness.

To measure the clean accuracy, we use all 10\,000 test images from the MNIST dataset.
The ptb-robustness is measured with 100 perturbation-based adversarial examples based on the MNIST test samples, crafted with the projected gradient descent attack ($\epsilon =0.3$). To measure the inv-robustness we use the 100 human-crafted examples provided by Tramer et al.~\cite{pmlr-v119-tramer20a}. 
Similar to Tramer et al.~\cite{pmlr-v119-tramer20a}, who used human participants to classify the testing data, we count an invariance-based adversarial example as correctly classified if the classification of the CNN matches the human-defined label.

\section{Experimental Results}\label{sec:results}
\begin{figure}
    \centering
    \begin{subfigure}[b]{0.45\textwidth}
        \includegraphics[width=\textwidth]{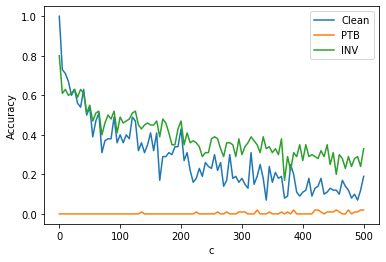}
        \caption{standard-trained and AL}
        \label{fig:a}
    \end{subfigure}
    \quad
    \begin{subfigure}[b]{0.45\textwidth}
        \includegraphics[width=\textwidth]{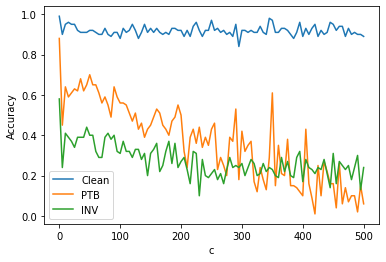}
        \caption{ptb-trained and AL}
        \label{fig:b}
    \end{subfigure}
    
    \vspace{3mm}
    \begin{subfigure}[b]{0.45\textwidth}
        \includegraphics[width=\textwidth]{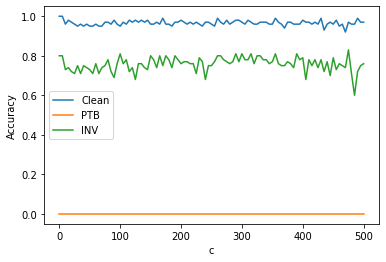}
        \caption{standard-trained and HL}
        \label{fig:c}
    \end{subfigure}
    \quad
    \begin{subfigure}[b]{0.45\textwidth}
        \includegraphics[width=\textwidth]{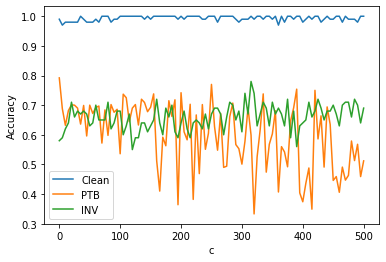}
        \caption{ptb-trained and HL}
        \label{fig:d}
    \end{subfigure}
    %\vspace{-3mm}
    \caption{Comparison of test accuracies during retraining using invaraiance-based adversarial examples (inv-based training) across different models and data labeling methods. AL indicates algorithm-defined training data labels, HL stands for human-defined training data labels. For all Figures, the x-axis represents the number of invariance-based adversarial examples. The y-axis depicts the accuracy and robustness values with respect to benign data~(Clean), perturbation-based adversarial examples~(PTB), and invariance-based adversarial examples~(INV).
    }
    \label{fig:results}
\end{figure}
The accuracy and robustness values of the standard-trained and ptb-trained model are listed in the top two rows of Table~\ref{tab:results}.
Clean accuracy describes the accuracy on the MNIST test data (clean acc.), ptb-robustness (ptb-rob.) describes the accuracy on the perturbation-based adversarial examples and inv-robustness (inv-rob.) describes the accuracy on the invariance-based testing data provided by Tramer et al.~\cite{pmlr-v119-tramer20a}.

It is clearly visible that with rising ptb-robustness, the inv-robustness is decreasing, confirming the findings of~\cite{jacobsen2019exploiting}. The standard-trained model has an inv-robustness of 80.0\,\% and when the perturbation-based adversarial training is finished, the inv-robustness decreases to 57.9\,\%. Clean accuracy on the MNIST test data decreases from 99.9\,\% to 96.2\,\%, so the perturbation-based adversarial training only has a minor negative effect on it.
While the standard-trained model exhibits low ptb-robustness, the ptb-trained model has only limited robustness against invariance-based adversarial examples.
Hence, none of the models achieved high ptb-robustness and high inv-robustness using perturbation-based adversarial training.
%
%\subsection{Influence of the labels on invariance-based adversarial training}
\\\\
When performing adversarial training on the standard-trained and ptb-trained model using invariance-based adversarial examples, accuracy and robustness values evolve as illustrated in Figure~\ref{fig:results}.
First, the results for training the standard-trained model using invariance-based adversarial examples with algorithm-defined labels results in accuracy and robustness values as depicted in Figure~\ref{fig:a}. It is clearly visible that clear accuracy and inv-robustness both decrease using the algorithm-defined labels.
Figure~\ref{fig:b} shows the results of invariance-based adversarial training based on the ptb-trained model and the algorithm-defined labels.
In contrast to the standard-trained model, here clean accuracy is not decreasing, however ptb-robustness is significantly reduced.
Next, Figure~\ref{fig:c} illustrates the results for retraining the standard-trained model using invariance-based adversarial examples with human-defined labels. Clean accuracy and inv-robustness slightly decrease while ptb-robustness remains constant at zero.
Lastly, the values during adversarial training with the ptb-based model using invariance-based adversarial examples with human-defined labels can be seen in Figure~\ref{fig:d}. 
Clean accuracy and inv-robustness remain on a relatively constant level, but ptb-robustness shows strong fluctuations with a descending trend.
Lines 3--6 in Table~\ref{tab:results} summarize the accuracy and robustness values at the end of all four training runs ($c=500$).

In conclusion, no experiment showed a clear positive effect on accuracy and robustness of the model. Thus, it is generally not recommended to perform invariance-based adversarial training in this form.
Considering which labels to use when performing invariance-based adversarial training, our experiments showed that human-defined labels are the better choice.

\begin{table}[]
    \centering
    \caption{Accuracy and robustness values for the examined models. AL indicates algorithm-defined training data labels, HL stands for human-defined training data labels. The term on the left of the arrow indicates the base model's training procedure, the term on the right indicates the additional training procedure}
    \begin{tabular}{c R{3.2cm} c L{3.2cm} c c c}
        \toprule
        & & & & clean acc. & ptb-rob. & inv-rob. \\
        \midrule
        & \multicolumn{3}{c}{standard-trained} & 99.9\,\% & \phantom{0}0.0\,\% & 80.0\,\% \\
        & \multicolumn{3}{c}{ptb-trained} & 96.2\,\% & 88.9\,\% & 57.9\,\% \\
        \midrule
        \multirow{2}{*}{AL} & standard-trained & $\rightarrow$ & inv-based training & \phantom{0}7.9\,\% & \phantom{0}7.8\,\% & 25.9\,\% \\
        & ptb-trained & $\rightarrow$ & inv-based training & 92.0\,\% & 17.4\,\% & 28.0\,\% \\
        \midrule
        \multirow{3}{*}{HL} & standard-trained & $\rightarrow$ & inv-based training & 95.0\,\% & \phantom{0}0.0\,\% & 77.9\,\% \\
        & ptb-trained & $\rightarrow$ & inv-based training & 99.0\,\% & 52.1\,\% & 67.0\,\% \\[2ex]
        & standard-trained & $\rightarrow$ & simultaneous training & 96.0\,\% & 79.0\,\% & 71.0\,\% \\
        % \midrule
        % & standard-trained & $\rightarrow$ & simu. training (923) & 100 & 87.0 & 76.0 \\
        % & standard-trained & $\rightarrow$ & simu. training (1227) & 99.0 & 89.0 & 71.0 \\
        % & standard-trained & $\rightarrow$ & simu. training (1281) & 99.0 & 88.0 & 61.0 \\
        % & standard-trained & $\rightarrow$ & simu. training (1291) & 99.0 & 86.0 & 70.0 \\
        % & standard-trained & $\rightarrow$ & simu. training (1338) & 100 & 82.0 & 74.0 \\
        % \midrule
        % & 3-rolling-avg & & 929 -- 931 & 99.0 & 81.0 & 77.3 \\
        % & 3-rolling-avg & & 938 -- 940 & 99.3 & 80.7 & 77.3 \\
        % & 3-rolling-avg & & 981 -- 983 & 100 & 85.3 & 71.3 \\
        % & 3-rolling-avg & & 1040 -- 1042 & 97.0 & 84.0 & 74.0 \\
        \bottomrule
    \end{tabular}
    \label{tab:results}
\end{table}
% \begin{wrapfigure}{r}{.5\textwidth}
%     \centering
%     \includegraphics[scale=0.5]{img/output1.png}
%     \captionof{figure}{Accuracy (robustness) for benign data~(Clean), perturbation-based adversarial examples~(PTB), and invariance-based adversarial examples~(INV) during adversarial retraining using perturbation-based and invariance-based adversarial examples simultaneously.}
%     \label{fig:simultan-ptb-inv-training}
% \end{wrapfigure}

\section{Simultaneous Adversarial Training}\label{sec:simultaneous}
When performing consecutive ptb-based and inv-based adversarial training, the resulting models did not improve both, accuracy and robustness. Thus, we further investigate a simultaneous training set-up on both, perturbation-based and invariance-based adversarial examples.
%We perform both adversarial training approaches simultaneously.
In each iteration of Algorithm~\ref{alg:ptb-training}, besides the perturbation-based adversarial examples, we also include the 500 invariance-based adversarial examples with the human-defined labels. We assume, that this might result in an improved inv-robustness and ptb-robustness.

Figure~\ref{fig:simultan_train} shows the accuracy and robustness values during adversarial retraining when including invariance-based adversarial examples.
This kind of adversarial training has only a negligible negative effect on the clean accuracy. The inv-robustness goes down from 80.0\,\% to 71.0\,\% and the ptb-robustness goes up from 0.0\,\% to 79.0\,\%. This means that we are able to achieve a ptb-robustness of 79.0\,\% simultaneously to an inv-robustness of 71.0\,\%.

Adversarial training using perturbation-based and invariance-based adversarial examples simultaneously does not result in a model outperforming the ptb-trained model considering robustness against both types of adversarial examples.
However, when taking the ratio between perturbation-based and invariance-based adversarial examples into account, we can see that the simultaneously trained model outperforms all other models in terms of ``combined robustness'' in a large range, as illustrated in Figure~\ref{fig:tradeoff}. The combined robustness represents a mixture of the robustnesses against invariance-based and perturbation-based adversarial examples depending on the ratio $\psi$ between invariance-based and perturbation-based examples
which is defined as:
\begin{equation}
    \psi \coloneqq \frac{\text{\# inv-based adv. ex.}}{\text{\# inv-based adv. ex.} + \text{\# ptb-based adv. ex.}}\,.
\end{equation}
For example, $\psi = 0$ indicates that a model owner faces only perturbation-based adversarial examples, while $\psi = 1$ indicates that only invariance-based adversarial examples are present\footnote{Note that we expect an adversary to attack 100\,\% of all test samples, a scenario that is challenged by recent research~\cite{samsinger2021should}.}.
Consequently, the combined robustness of a model for a certain ratio $\psi$ can be calculated by $\psi\cdot\text{inv-robustness} + (1-\psi)\cdot\text{ptb-robustness}$.

\section{Discussion}\label{sec:discussion}
While standard-training and ptb-adversarial training achieve good robustness against either invariance-based or perturbation-based adversarial examples, they have the huge downside of having limited to none robustness against the respective other type of adversarial examples, posing a natural trade-off for model owners that need to defend against both adversarial example types. %As the type of adversarial examples a model owner is confronted with is likely unknown. 
We show that simultaneous adversarial training on perturbation-based and invariance-based adversarial examples results in the best accuracy for a wide range of $\psi$-values (.430 to .898), thus providing the model-owner with an additional, competitive option to shape this trade-off.

Further, we experience instability in our training method, as visible in Figure~\ref{fig:simultan_train}. This allows to further optimize the procedure by implementing an early stopping mechanism (exemplary, we achieve 87\,\% ptb-robustness and 76\,\% inv-robustness after 923 iterations of our simultaneous training procedure). Future work should tackle this problem to achieve a smoother result space. 
% Summary of results
%We achieve the best accuracy and robustness values when simultaneously using perturbation-based and invariance-based adversarial examples instead of performing retraining using the two datasets one after the other.
% Interpretation
%Thus, our experiments show that adversarial training should be done using both types of adversarial examples simultaneously.
% Implication
% Limitation & Future Work
Further, our findings should be confirmed by experiments using other datasets and more complex networks in future work.
Our analysis incorporated invariance-based adversarial examples generated by one specific algorithm. Moreover, the number of adversarial examples used for retraining remains at a relatively low level of 500. Future work could build on our findings and perform experiments with larger datasets of adversarial examples, different network architectures, and varying $\epsilon$, towards an exhaustive analysis.

Finally, our work confirms and sheds more light into the fundamental trade-off between a model's invariance and sensitivity found by~\cite{pmlr-v119-tramer20a}, a problem widely disregarded in current research. As such, a sensitive first step would be to report the robustness against invariance-based adversarial examples in future research on defenses against adversarial examples.
\begin{figure}
\centering
\begin{minipage}[t]{.47\textwidth}
    \centering
    \scalebox{.8}{\input{simultan}}
    \captionof{figure}{Accuracies for benign data~(clean accuracy), perturbation-based adversarial examples~(ptb-robustness), and invariance-based adversarial examples~(inv-robustness) during adversarial retraining using perturbation-based and invariance-based adversarial examples simultaneously.}
    \label{fig:simultan_train}
\end{minipage}%
\qquad
\begin{minipage}[t]{.47\textwidth}
    \centering
    \scalebox{.8}{% This file was created by tikzplotlib v0.9.6.
\begin{tikzpicture}%[spy using outlines={circle, magnification=4, size=2cm, every spy on node/.append style={thick},connect spies}]

\definecolor{color0}{rgb}{0.12156862745098,0.466666666666667,0.705882352941177}
\definecolor{color1}{rgb}{1,0.498039215686275,0.0549019607843137}
\definecolor{color2}{rgb}{0.172549019607843,0.627450980392157,0.172549019607843}

\begin{axis}[
%width=\columnwidth,
%height=0.61803398875\columnwidth,
legend cell align={left},
legend style={fill opacity=0.8, draw opacity=1, text opacity=1, at={(0.97,0.1)}, anchor=south east, draw=white!80!black, nodes={scale=0.85, transform shape}},
tick align=outside,
tick pos=left,
x grid style={white!69.0196078431373!black},
xlabel={$\psi$},
xmajorgrids,
xmin=-0.05, xmax=1.05,
xtick style={color=black},
y grid style={white!69.0196078431373!black},
ylabel={Combined robustness},
ymajorgrids,
ymin=-0.04, ymax=1.04,
ytick style={color=black}
]
\addplot [thick, color0]
table {%
0 0.889
1 0.579
};
\addlegendentry{ptb-trained}
\addplot [thick, color1]
table {%
0 0
1 0.8
};
\addlegendentry{standard-trained}
\addplot [thick, color2]
table {%
0 0.79
1 0.71
};
\addlegendentry{simultaneously trained}

\addplot [dashed, black]
table {%
.43 0
.43 1
};

\addplot [dashed, black]
table {%
.898 0
.898 1
};

% absolute coordinate of area to be zoomed in
%\coordinate (SpyArea1) at (axis cs: 0.13,0.67);
% absolute coordinate of area showing zoomed
%\coordinate (SpyZoom1) at (axis cs: 0.4,0.8);

%\coordinate (SpyArea2) at (axis cs: 0.49,0.55);
% absolute coordinate of area showing zoomed
%\coordinate (SpyZoom2) at (axis cs: 0.8,0.8);
\end{axis}
%\spy on (SpyArea1) in node[fill=white] at (SpyZoom1);
%\spy on (SpyArea2) in node[fill=white] at (SpyZoom2);
\end{tikzpicture}}
    \captionof{figure}{Model accuracy for the dominant training procedures, the dashed lines indicate the points of intersection ($\psi=0.430$ and $\psi=0.898$).}
    \label{fig:tradeoff}
\end{minipage}
\end{figure}

\section{Conclusion}\label{sec:conclusion}
Adversarial training using perturbation-based adversarial examples has been investigated intensively over the past years. However, only little about the utilization of invariance-based adversarial examples for adversarial training is known.
In this paper, we show how the addition of such samples during training affect benign accuracy and robustness against perturbation-based and invariance-based adversarial examples of a CNN used for handwritten digit classification.

In conclusion, we can confirm a trade-off between the accuracy, perturbation-based robustness, and invariance-based robustness of ML models depending on the training procedure.
Furthermore, we identify sweet spots where simultaneous adversarial training, i.e. using perturbation-based and invariance-based adversarial examples at the same time, outperforms other techniques in terms of combined robustness against both, perturbation-based and invariance-based adversarial examples.

\begin{acks}
The second author is supported under the project ``Secure Machine Learning Applications with Homomorphically Encrypted Data'' (project no. 886524) by the Federal Ministry for Climate Action, Environment, Energy, Mobility, Innovation and Technology (BMK) of Austria. The third and fourth author are supported by the Austrian Science Fund (FWF) under grant no. I 4057-N31 (``Game Over Eva(sion)'').
\end{acks}

\bibliographystyle{ACM-Reference-Format}
\bibliography{references}


\begin{thebibliography}{20}

%%% the Web address.

\ifx \showCODEN    \undefined \def \showCODEN     #1{\unskip}     \fi
\ifx \showDOI      \undefined \def \showDOI       #1{#1}\fi
\ifx \showISBNx    \undefined \def \showISBNx     #1{\unskip}     \fi
\ifx \showISBNxiii \undefined \def \showISBNxiii  #1{\unskip}     \fi
\ifx \showISSN     \undefined \def \showISSN      #1{\unskip}     \fi
\ifx \showLCCN     \undefined \def \showLCCN      #1{\unskip}     \fi
\ifx \shownote     \undefined \def \shownote      #1{#1}          \fi
\ifx \showarticletitle \undefined \def \showarticletitle #1{#1}   \fi
\ifx \showURL      \undefined \def \showURL       {\relax}        \fi
% invisible to TeX
\providecommand\bibfield[2]{#2}
\providecommand\bibinfo[2]{#2}
\providecommand\natexlab[1]{#1}
\providecommand\showeprint[2][]{arXiv:#2}

\bibitem[Alazab and Tang(2019)]%
        {alazab2019deep}
\bibfield{author}{\bibinfo{person}{Mamoun Alazab} {and}
  \bibinfo{person}{MingJian Tang}.} \bibinfo{year}{2019}\natexlab{}.
\newblock \bibinfo{booktitle}{\emph{Deep learning applications for cyber
  security}}.
\newblock \bibinfo{publisher}{Springer}.
\newblock


\bibitem[Carlini and Wagner(2018)]%
        {carlini2018audio}
\bibfield{author}{\bibinfo{person}{Nicholas Carlini} {and}
  \bibinfo{person}{David Wagner}.} \bibinfo{year}{2018}\natexlab{}.
\newblock \showarticletitle{Audio Adversarial Examples: Targeted Attacks on
  Speech-to-Text}. In \bibinfo{booktitle}{\emph{2018 IEEE Security and Privacy
  Workshops (SPW)}}. IEEE, \bibinfo{pages}{1--7}.
\newblock


\bibitem[Dalvi et~al\mbox{.}(2004)]%
        {dalvi2004adversarial}
\bibfield{author}{\bibinfo{person}{Nilesh Dalvi}, \bibinfo{person}{Pedro
  Domingos}, \bibinfo{person}{Sumit Sanghai}, {and} \bibinfo{person}{Deepak
  Verma}.} \bibinfo{year}{2004}\natexlab{}.
\newblock \showarticletitle{Adversarial Classification}. In
  \bibinfo{booktitle}{\emph{Proceedings of the tenth ACM SIGKDD International
  Conference on Knowledge Discovery and Data Mining}}.
  \bibinfo{pages}{99--108}.
\newblock


\bibitem[Eykholt et~al\mbox{.}(2018)]%
        {eykholt2018robust}
\bibfield{author}{\bibinfo{person}{Kevin Eykholt}, \bibinfo{person}{Ivan
  Evtimov}, \bibinfo{person}{Earlence Fernandes}, \bibinfo{person}{Bo Li},
  \bibinfo{person}{Amir Rahmati}, \bibinfo{person}{Chaowei Xiao},
  \bibinfo{person}{Atul Prakash}, \bibinfo{person}{Tadayoshi Kohno}, {and}
  \bibinfo{person}{Dawn Song}.} \bibinfo{year}{2018}\natexlab{}.
\newblock \showarticletitle{Robust Physical-world Attacks on Deep Learning
  Visual Classification}. In \bibinfo{booktitle}{\emph{Proceedings of the IEEE
  Conference on Computer Vision and Pattern Recognition}}.
  \bibinfo{pages}{1625--1634}.
\newblock


\bibitem[Goodfellow et~al\mbox{.}(2015)]%
        {goodfellow2015explaining}
\bibfield{author}{\bibinfo{person}{Ian~J. Goodfellow},
  \bibinfo{person}{Jonathon Shlens}, {and} \bibinfo{person}{Christian
  Szegedy}.} \bibinfo{year}{2015}\natexlab{}.
\newblock \bibinfo{title}{Explaining and Harnessing Adversarial Examples}.
\newblock
\newblock
\showeprint[arxiv]{1412.6572}~[stat.ML]


\bibitem[Hinton et~al\mbox{.}(2012)]%
        {hinton2012neural}
\bibfield{author}{\bibinfo{person}{Geoffrey Hinton}, \bibinfo{person}{Nitish
  Srivastava}, {and} \bibinfo{person}{Kevin Swersky}.}
  \bibinfo{year}{2012}\natexlab{}.
\newblock \showarticletitle{Neural networks for machine learning lecture 6a
  overview of mini-batch gradient descent}.
\newblock \bibinfo{journal}{\emph{Cited on}} \bibinfo{volume}{14},
  \bibinfo{number}{8} (\bibinfo{year}{2012}), \bibinfo{pages}{2}.
\newblock


\bibitem[Huang et~al\mbox{.}(2020)]%
        {huang2020deep}
\bibfield{author}{\bibinfo{person}{Jian Huang}, \bibinfo{person}{Junyi Chai},
  {and} \bibinfo{person}{Stella Cho}.} \bibinfo{year}{2020}\natexlab{}.
\newblock \showarticletitle{Deep learning in finance and banking: A literature
  review and classification}.
\newblock \bibinfo{journal}{\emph{Frontiers of Business Research in China}}
  \bibinfo{volume}{14} (\bibinfo{year}{2020}), \bibinfo{pages}{1--24}.
\newblock


\bibitem[Jacobsen et~al\mbox{.}(2020)]%
        {jacobsen2020excessive}
\bibfield{author}{\bibinfo{person}{J{\"o}rn-Henrik Jacobsen},
  \bibinfo{person}{Jens Behrmann}, \bibinfo{person}{Richard Zemel}, {and}
  \bibinfo{person}{Matthias Bethge}.} \bibinfo{year}{2020}\natexlab{}.
\newblock \bibinfo{title}{Excessive Invariance Causes Adversarial
  Vulnerability}.
\newblock
\newblock
\showeprint[arxiv]{1811.00401}~[cs.LG]


\bibitem[Jacobsen et~al\mbox{.}(2019)]%
        {jacobsen2019exploiting}
\bibfield{author}{\bibinfo{person}{J{\"o}rn-Henrik Jacobsen},
  \bibinfo{person}{Jens Behrmannn}, \bibinfo{person}{Nicholas Carlini},
  \bibinfo{person}{Florian Tram{\`e}r}, {and} \bibinfo{person}{Nicolas
  Papernot}.} \bibinfo{year}{2019}\natexlab{}.
\newblock \bibinfo{title}{Exploiting Excessive Invariance Caused by
  Norm-Bounded Adversarial Robustness}.
\newblock
\newblock
\showeprint[arxiv]{1903.10484}~[cs.LG]


\bibitem[Ker et~al\mbox{.}(2017)]%
        {ker2017deep}
\bibfield{author}{\bibinfo{person}{Justin Ker}, \bibinfo{person}{Lipo Wang},
  \bibinfo{person}{Jai Rao}, {and} \bibinfo{person}{Tchoyoson Lim}.}
  \bibinfo{year}{2017}\natexlab{}.
\newblock \showarticletitle{Deep learning applications in medical image
  analysis}.
\newblock \bibinfo{journal}{\emph{Ieee Access}}  \bibinfo{volume}{6}
  (\bibinfo{year}{2017}), \bibinfo{pages}{9375--9389}.
\newblock


\bibitem[LeCun et~al\mbox{.}(2010)]%
        {lecun2010mnist}
\bibfield{author}{\bibinfo{person}{Yann LeCun}, \bibinfo{person}{Corinna
  Cortes}, {and} \bibinfo{person}{Chris Burges}.}
  \bibinfo{year}{2010}\natexlab{}.
\newblock \bibinfo{title}{M{NIST} handwritten digit database}.
\newblock
\newblock
\urldef\tempurl%
\url{http://yann.lecun.com/exdb/mnist/}
\showURL{%
\tempurl}


\bibitem[Madry et~al\mbox{.}(2018)]%
        {madry2018towards}
\bibfield{author}{\bibinfo{person}{Aleksander Madry},
  \bibinfo{person}{Aleksandar Makelov}, \bibinfo{person}{Ludwig Schmidt},
  \bibinfo{person}{Dimitris Tsipras}, {and} \bibinfo{person}{Adrian Vladu}.}
  \bibinfo{year}{2018}\natexlab{}.
\newblock \showarticletitle{Towards Deep Learning Models Resistant to
  Adversarial Attacks}. In \bibinfo{booktitle}{\emph{International Conference
  on Learning Representations}}.
\newblock


\bibitem[Metzen et~al\mbox{.}(2017)]%
        {metzen2017detecting}
\bibfield{author}{\bibinfo{person}{Jan~Hendrik Metzen}, \bibinfo{person}{Tim
  Genewein}, \bibinfo{person}{Volker Fischer}, {and} \bibinfo{person}{Bastian
  Bischoff}.} \bibinfo{year}{2017}\natexlab{}.
\newblock \bibinfo{title}{On Detecting Adversarial Perturbations}.
\newblock
\newblock
\showeprint[arxiv]{1702.04267}~[stat.ML]


\bibitem[Pang et~al\mbox{.}(2018)]%
        {pang2018robust}
\bibfield{author}{\bibinfo{person}{Tianyu Pang}, \bibinfo{person}{Chao Du},
  \bibinfo{person}{Yinpeng Dong}, {and} \bibinfo{person}{Jun Zhu}.}
  \bibinfo{year}{2018}\natexlab{}.
\newblock \bibinfo{title}{Towards Robust Detection of Adversarial Examples}.
\newblock
\newblock
\showeprint[arxiv]{1706.00633}~[cs.LG]


\bibitem[Rauber et~al\mbox{.}(2017)]%
        {rauber2017foolbox}
\bibfield{author}{\bibinfo{person}{Jonas Rauber}, \bibinfo{person}{Wieland
  Brendel}, {and} \bibinfo{person}{Matthias Bethge}.}
  \bibinfo{year}{2017}\natexlab{}.
\newblock \showarticletitle{Foolbox: {{A Python}} Toolbox to Benchmark the
  Robustness of Machine Learning Models}. In \bibinfo{booktitle}{\emph{Reliable
  Machine Learning in the Wild Workshop, 34th International Conference on
  Machine Learning}}.
\newblock


\bibitem[Rauber et~al\mbox{.}(2020)]%
        {rauber2017foolboxnative}
\bibfield{author}{\bibinfo{person}{Jonas Rauber}, \bibinfo{person}{Roland
  Zimmermann}, \bibinfo{person}{Matthias Bethge}, {and}
  \bibinfo{person}{Wieland Brendel}.} \bibinfo{year}{2020}\natexlab{}.
\newblock \showarticletitle{Foolbox {{Native}}: {{Fast}} Adversarial Attacks to
  Benchmark the Robustness of Machine Learning Models in {{PyTorch}},
  {{TensorFlow}}, and {{JAX}}}.
\newblock \bibinfo{journal}{\emph{Journal of Open Source Software}}
  \bibinfo{volume}{5}, \bibinfo{number}{53} (\bibinfo{year}{2020}),
  \bibinfo{pages}{2607}.
\newblock
\urldef\tempurl%
\url{https://doi.org/10.21105/joss.02607}
\showDOI{\tempurl}


\bibitem[Samsinger et~al\mbox{.}(2021)]%
        {samsinger2021should}
\bibfield{author}{\bibinfo{person}{Maximilian Samsinger},
  \bibinfo{person}{Florian Merkle}, \bibinfo{person}{Pascal Sch{\"o}ttle},
  {and} \bibinfo{person}{Tomas Pevny}.} \bibinfo{year}{2021}\natexlab{}.
\newblock \showarticletitle{When Should You Defend Your Classifier? -- A
  Game-Theoretical Analysis of Countermeasures Against Adversarial Examples}.
  In \bibinfo{booktitle}{\emph{Decision and Game Theory for Security: 12th
  International Conference, GameSec 2021, Virtual Event, October 25--27, 2021,
  Proceedings}}. Springer, \bibinfo{pages}{158--177}.
\newblock


\bibitem[Sharif et~al\mbox{.}(2016)]%
        {sharif2016accessorize}
\bibfield{author}{\bibinfo{person}{Mahmood Sharif}, \bibinfo{person}{Sruti
  Bhagavatula}, \bibinfo{person}{Lujo Bauer}, {and} \bibinfo{person}{Michael~K
  Reiter}.} \bibinfo{year}{2016}\natexlab{}.
\newblock \showarticletitle{Accessorize to a crime: Real and stealthy attacks
  on state-of-the-art face recognition}. In
  \bibinfo{booktitle}{\emph{Proceedings of the 2016 ACM Sigsac Conference on
  Computer and Communications Security}}. \bibinfo{pages}{1528--1540}.
\newblock


\bibitem[Szegedy et~al\mbox{.}(2014)]%
        {szegedy2013intriguing}
\bibfield{author}{\bibinfo{person}{Christian Szegedy},
  \bibinfo{person}{Wojciech Zaremba}, \bibinfo{person}{Ilya Sutskever},
  \bibinfo{person}{Joan Bruna}, \bibinfo{person}{Dumitru Erhan},
  \bibinfo{person}{Ian Goodfellow}, {and} \bibinfo{person}{Rob Fergus}.}
  \bibinfo{year}{2014}\natexlab{}.
\newblock \showarticletitle{Intriguing properties of neural networks}. In
  \bibinfo{booktitle}{\emph{International Conference on Learning
  Representations}}.
\newblock


\bibitem[Tramer et~al\mbox{.}(2020)]%
        {pmlr-v119-tramer20a}
\bibfield{author}{\bibinfo{person}{Florian Tramer}, \bibinfo{person}{Jens
  Behrmann}, \bibinfo{person}{Nicholas Carlini}, \bibinfo{person}{Nicolas
  Papernot}, {and} \bibinfo{person}{Joern-Henrik Jacobsen}.}
  \bibinfo{year}{2020}\natexlab{}.
\newblock \showarticletitle{Fundamental Tradeoffs between Invariance and
  Sensitivity to Adversarial Perturbations}. In
  \bibinfo{booktitle}{\emph{Proceedings of the 37th International Conference on
  Machine Learning}} \emph{(\bibinfo{series}{Proceedings of Machine Learning
  Research}, Vol.~\bibinfo{volume}{119})},
  \bibfield{editor}{\bibinfo{person}{Hal~Daum{\'e} III} {and}
  \bibinfo{person}{Aarti Singh}} (Eds.). \bibinfo{publisher}{{PMLR}},
  \bibinfo{pages}{9561--9571}.
\newblock


\end{thebibliography}

% Not sure if we need an appendix
%\appendix

\end{document}